%% file: main.tex
\documentclass{article}

\PassOptionsToPackage{numbers, compress}{natbib}

\usepackage[preprint]{neurips_2023}

\usepackage[utf8]{inputenc} 
\usepackage[T1]{fontenc}    
\usepackage{hyperref}       
\usepackage{url}            
\usepackage{booktabs}       
\usepackage{amsfonts}       
\usepackage{nicefrac}       
\usepackage{microtype}      
\usepackage{xcolor}         
\usepackage{graphicx}
\usepackage{amsmath}

\def\method{Geneverse}

\title{Geneverse: A collection of Open-source Multimodal Large Language Models for Genomic and Proteomic Research}

\author{%
  Tianyu Liu \\
  Yale University\\
  New Haven, CT 06511 \\
  \texttt{tianyu.liu@yale.edu} \\
  \And
  Yijia Xiao \\
  UCLA \\
  Los Angeles, CA 90095 \\
  \texttt{yijia.xiao@cs.ucla.edu} \\
  \And
  Xiao Luo \\
  UCLA \\
  Los Angeles, CA 90095 \\
  \texttt{xiao.luo@cs.ucla.edu} \\
  \And
  Hua Xu \\
  Yale University \\
  New Haven, CT 06511 \\
  \texttt{hua.xu@yale.edu} \\
  \And
  W. Jim Zheng \\
  UT Health \\
  Houston, TX 77030 \\
  \texttt{Wenjin.J.Zheng@uth.tmc.edu} \\
  \And
  Hongyu Zhao \\
  Yale University \\
  New Haven, CT 06511 \\
  \texttt{hongyu.zhao@yale.edu} \\
}

\begin{document}
\maketitle
\begin{abstract}
The applications of large language models (LLMs) are promising for biomedical and healthcare research. Despite the availability of open-source LLMs trained using a wide range of biomedical data, current research on the applications of LLMs to genomics and proteomics is still limited. To fill this gap, we propose a collection of finetuned LLMs and multimodal LLMs (MLLMs), known as Geneverse, for three novel tasks in genomic and proteomic research. The models in Geneverse are trained and evaluated based on domain-specific datasets, and we use advanced parameter-efficient finetuning techniques to achieve the model adaptation for tasks including the generation of descriptions for gene functions, protein function inference from its structure, and marker gene selection from spatial transcriptomic data. We demonstrate that adapted LLMs and MLLMs perform well for these tasks and may outperform closed-source large-scale models based on our evaluations focusing on both truthfulness and structural correctness. All of the training strategies and base models we used are freely accessible.
\end{abstract}

\section{Introduction}

Foundation Models (FMs) \citep{bommasani2021opportunities} have attracted great attention recently because of their superb functionality for handling multiple downstream tasks by adaption, especially in the landscape of Natural Language Processing (NLP) \citep{wu2023survey, zhao2023survey}. In this area, we focus on one specific type of FMs, known as generative Large Language Models (LLMs) \citep{zhao2023survey}. LLMs have demonstrated their strong ability to perform various tasks and domains in NLP \citep{bommasani2021opportunities, zhao2023survey}, and they have revolutionized the approaches of productivity improvement with machine intelligence \citep{byun2023dispensing} as well. Powerful LLMs including ChatGPT \citep{achiam2023gpt} and Bard \citep{team2023gemini} exhibit the ability to understand and reason as a human, and they are capable of complicated problem-solving tasks. However, since these well-trained models are both closed-source, the emergence of open-source LLM development is also important. Powerful open-source LLMs including LLaMA2 \citep{touvron2023llama} and Mistral \citep{jiang2023mistral} have already become the base models of many LLMs for scientific research \citep{labrak2024biomistral, wu2023pmc}.

Beyond the success of LLMs, we are also interested in extending the input of LLMs with datasets from different modalities, including images \citep{liu2024visual} and scientific graphs \citep{wei2023unleashing}. Such extensions of LLMs are meaningful because different modalities offer different aspects of a good representation \citep{wu2023multimodal, huh2024platonic}. Therefore, Multimodal LLMs (MLLMs) are emerging as a new research hotspot \citep{wu2023next}. Models like Gemini (closed-source) \citep{team2023gemini}, GPT-4 (closed-source) \citep{achiam2023gpt} and LLaVA (open-source) \citep{liu2024visual} allow the combination of images and texts as input, thus making MLLMs understand non-text data and generate responses accordingly \citep{liu2023radiology}. The performance of these models demonstrates that treating images as a special language can extend the capabilities of LLMs to solve more tasks including image-based conversation, image description and complex reasoning \citep{li2024llava, mesko2023impact}. 

Although we have seen successful examples related to the development of LLMs and MLLMs, the application of LLMs towards healthcare and biomedical research presents both challenges and opportunities \citep{he2023survey, zhou2023survey}. First, different from other areas, biomedical research highlights strict requirements for the factualness or truthfulness of model outputs \citep{labrak2024biomistral}, as well as validation of outputs by clinical or biological experiments \citep{thirunavukarasu2023large}. When LLMs are used to assist physicians in the diagnosis of patients, an incorrect diagnosis of a patient's condition can dramatically affect the process and even cause the patient to miss the optimal time for treatment. Second, the output of LLMs should also comply with established scientific knowledge \citep{arora2023theory}. Third, the performance of finetuned open-source LLMs also has not outperformed large or proprietary models \citep{labrak2024biomistral} for such research. Regarding the opportunities, most of the LLMs for biomedical research focus on precision healthcare \citep{wu2023pmc} and are finetuned with information from the corpus of biomedical articles \citep{luo2023biomedgpt}, while the LLMs or MLLMs for genomic or proteomic research are not well-discussed and remain to be explored. Therefore, we aim to contribute such FMs to accelerate research related to Central Dogma \citep{shapiro2009revisiting} at the molecular level \citep{brownlee2023enhancing}.

To fill the gap in this area, we propose a collection of LLMs and MLLMs, named as \method{} \footnote[1]{The name is inspired by the combination of GeneGPT and universe.}, for solving genomic and proteomic problems. Our collection mainly focuses on the functions of genes and proteins, as well as their interactions. Overall, our contributions can be summarized as follows:

1. By performing a comprehensive benchmarking analysis for different base models finetuned with the gene functional description data, we created a set of LLMs with different scales for genomic research. Our benchmark results included evaluations for both the correctness of structure and the truthfulness. We selected the best base models and created an Artificial Intelligence (AI) assistant.

2. By leveraging the training datasets from both official databases as well as synthetic data from advanced LLMs, we proposed a new adaptation framework for biomedical research and built a set of powerful LLMs based on the best base model for generating the summary of gene functions and protein functions after benchmarking. 

3. We introduced MLLMs in genomic and proteomic research by incorporating the structural information from proteins and expression information from genes, as well as their functional descriptions, as input. We then finetuned MLLMs for two tasks including protein identification and marker gene identification.

4. We explored the possibility of handling more tasks with finetuned FMs as well as improving the performances of models with more advanced techniques, which offered guidance for future biomedical AI research.

\begin{figure*}[ht]
    \centering
    \includegraphics[width=0.87\textwidth]{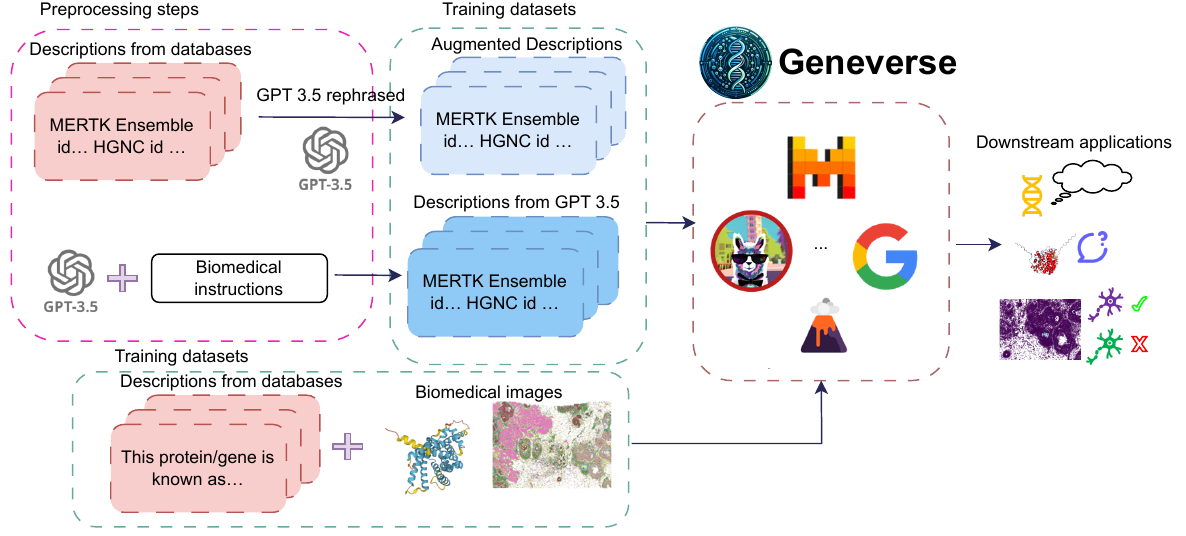}
    \caption{The landscape of \method{}. To generate LLMs for genomic and proteomic analysis, we incorporate the training datasets from rephrased descriptions for gene functions as well as synthetic descriptions from GPT 3.5. We then adjust the base model with different strategies and select the best candidate. To generate MLLMs for genomic and proteomic analysis, we incorporate the training datasets from known databases, including both descriptions and corresponding images. We then finetune the base model with different strategies and select the best candidate. The logo of \method{} is generated by DALLE \citep{dalle}.
    \label{fig:model overview}}
    \vspace{-0.3cm}
\end{figure*}

\section{Related Work}
\subsection{LLMs and MLLMs for general purpose}
General purpose models including GPT-4 \citep{achiam2023gpt}, Gemini \citep{team2023gemini} and Claude \citep{enis2024llm} have demonstrated exceptional performances across various tasks in language processing, under both zero-shot and few-shots learning frameworks. However, all of the models above are closed-source. Meanwhile, there are also attractive open-source LLMs including LLaMA series \citep{touvron2023llama, metallama2024}, Mistral \citep{jiang2023mistral} and Gemma \citep{gemma2024}, which also demonstrate comparable performances with proprietary models under certain tasks. Furthermore, the capacities of LLMs with multimodal information (e.g., audio, images, and DNA sequences, etc.) as inputs/outputs, known as MLLMs, have also shown remarkable performances under many multi-modal tasks. However, the models that serve as generalists often fail to answer certain domain-specific questions, for example, in the biological domain. 

\subsection{LLMs and MLLMs for biomedical research}
The adaptation of LLMs and MLLMs to address tasks in biology and medicine also attracts research communities. This topic studies how to leverage these models' abilities to address biomedical problems. Encoder-based models including BioBERT \citep{lee2020biobert} and MedBERT \citep{rasmy2021med} have been designed for answering biomedical questions. Recently, decoder-based models including BioGPT \citep{luo2022biogpt}, BioMedLM \citep{BioMedLM2023}, ClinicalGPT \citep{wang2023clinicalgpt}, BioMedGPT \citep{luo2023biomedgpt}, MEDITRON \citep{bosselut2024meditron}, BioMistral \citep{labrak2024biomistral} and Med-PALM 2 \citep{qian2024liver} are pre-trained or fine-tuned with biomedical databases and used for answering questions in genomics, proteomics, and clinical informatics, etc. Other methods including GeneGPT \citep{jin2023genegpt} and scELMo \citep{liu2023scelmo} utilize tool augmentation or prompt engineering to address similar questions without extra training. By jointly modelling biomedical images and texts, methods including LLaVA-Med \citep{li2024llava} and Med-MLLM \citep{liu2023medical}, have been designed to address multimodal problems. However, some of the models limit the accessibility while others do not show satisfactory performances in our targeted tasks. Therefore, we still need a better foundation model for healthcare and biology research.

\section{Methods}

\subsection{Problem statement} 
In this manuscript, we focus on an LLM or MLLM $\mathcal{M}$, which accepts the prompt $P(T,I,m)$ as input, where $T$ represents the text description, $I$ represents the image and $m$ represents the given task. For an LLM, we set $I$ as an empty item. For a MLLM, we use both $T$ and $I$. For each prompt, we have a corresponding model output as $O = \mathcal{M}(P(T,I,m))$. For a validation dataset, we compare the difference between $O$ and the ground truth information $G$ to evaluate model performance. The more similar $O$ and $G$ are (The higher their similarity score is), the better the model is.

\subsection{Overview of \method{}} 
Our model collection \method{} is based on a group of pre-training open-source foundation models including LLaMA2-7B, LLaMA2-13B, Mistral-7B, Gemma-7B \citep{gemma2024}, LLaMAPro-8B \citep{wu2024llama}, LLaMA3-8B \citep{metallama2024} and LLaVA-7B. The first six models are designed for one text-based task, while the last one is designed for two image-text-based tasks. We finetuned these models based on both Parameter-Efficient FineTuning (PEFT) technology and full-parameter finetuning technology, and the finetuning setting is consistent with the Supervised Instruct Finetuning (SIFT) approach. The PEFT technology we used is Low-rank Adaption (LoRA) \citep{hu2022lora}. Our training datasets are constructed based on augmented datasets from the National Center of Biotechnology Information (NCBI) \citep{sayers2021database} and the generated synthetic datasets from GPT 3.5 with designed instructions. In the inference process, we did not use the sampling mode and thus the outputs were not affected by random seeds. The training mechanism is summarized in Figure \ref{fig:model overview}. We include the details of problem ettings and training design in the following subsections.

\subsection{Supervised finetuning process of LLM} 
To finetune a LLM as an AI assistant for genomic and proteomic information queries, we prepare the training dataset with both real data and synthetic data with a data augmentation policy. To collect the real data, we use scripts to access the gene information from NCBI pages and rephrase the content of each gene with GPT 3.5 to normal sequence. To collect the synthetic data, we use the same prompt to ask GPT 3.5 to generate responses for protein-encoding genes. By combining two different datasets together, we finalize our finetuning datasets. During the finetuning process, we search the optimized hyper-parameter settings and follow the default evaluation settings of Stanford-alpaca \citep{stanfordalpaca} and Alpaca-lora \citep{alpacalora2023}.

\subsection{Supervised finetuning process of MLLM} 
To finetune a MLLM as a multimodal AI assistant for genomic and proteomic information query, we prepare the training datasets for two representative tasks in biology. Learning a protein's function based on its structure (the extra modality) and identifying marker genes for certain cell types based on the locations of cells (the extra modality) are difficult but meaningful \citep{bernhofer2021predictprotein, yuan2024benchmarking}. Therefore, we consider the protein function classification task (known as the protein task) for protein structures and the maker genes identification task for spatial transcriptomics (known as the marker gene task). For the protein task, we downloaded the protein structure images from the Deepmind Alphafold2 website \citep{jumper2021highly} with the fixed capture angle. We load the image information using Pymol \citep{yuan2017using} and construct the instruction tuning dataset. We then finetune LLaVA with LoRA based on our datasets. We search the optimized hyper-parameter settings and use the default evaluation settings from LLaVA v1.5. For the marker gene task, the pipeline is the same but we utilize different image datasets \citep{lin2020scclassify} as well as different instructions.

\subsection{Post-processing steps}
To tackle the two problems of the outputs of LLMs, known as the incorrect inference of numerical features of genes and redundant descriptions (also known as degeneration) of genes, we design two approaches as post-processing methods to improve the quality of model outputs.

Our post-processing methods are based on the idea of tool-augmented design \citep{li2023api}. To handle the first problem, we collect the gene ID information from public databases \citep{sayers2021database} and replace the information in the outputs of LLMs with the numerical information from the database. To handle the second problem, since the redundant information is related to the aliases of genes, we delete the content related to aliases in the model outputs and insert the aliases based on the correct information from our database.

\subsection{Evaluation} 
In the evaluation process, we consider two different aspects for evaluation, known as grader for truthfulness and grader for structural correctness. The former grader works for evaluating whether the outputs from LLMs or MLLMs match the fact, while the latter grader works for evaluating whether the outputs from LLMs or MLLMs match the structure requirements from the prompt. Higher scores mean better model performance for each grader. 

We first consider the grader for the truthfulness. For the gene description task, we assign 1 for the outputs of LLMs which match the major gene function and gene name, while 0 otherwise. For the protein task, we compute the length of the largest common string between the output of MLLMs and the correct result and divide it by the length of the correct result as the metric. For the marker gene task, we assign 1 for the outputs of MLLMs which match the ground truth marker gene-cell type relation, while 0 otherwise. We finally average the scores of different samples as the final score.

We then consider the grader for structural correctness. For all of the tasks we evaluate, we assign 1 for the outputs if they match the structural requirements from the prompt, and 0 otherwise. We finally average the scores of different samples as the final score.

In order to validate our proposed metric, we also considered the use of BiLingual Evaluation Understudy (BLEU) \citep{papineni2002bleu} and ROUGE$k$ (k Grams Recall-Oriented Understudy for Gisting Evaluation) \citep{lin2004rouge} scores as additional evaluations. BLEU score is based on the overlap value of $n-grams$ for two strings. ROUGE$k$ score considers the overlap of $k$ grams between the reference text and generated text, and here $k=1$. In order to evaluate the quality of embeddings from generated texts, we computed the Normalized Mutual Information (NMI) score \citep{pedregosa2011scikit} between Leiden clusters \citep{traag2019louvain} and observed labels. All metrics are in [0,1] and higher values mean better model performances. 

\section{Results}
\subsection{Benchmarking LLMs for summarizing of gene functions} 
We first performed benchmarking analysis by fixing the training datasets to find the best base model and the best training strategy. Our task is to generate the summary of gene functions based on the prompt only containing the task description. Out of $\sim$ 20,000 genes \citep{dolgin2017most} in total, we randomly selected 20 genes referred from \citep{jin2023genegpt} and their observed descriptions to construct the evaluation datasets. Since our prompt also contains requirements for the structure of model outputs, we consider a novel evaluation of the output structure as well as the truthfulness of the contents. The illustration of evaluators is shown in Extended Data Figure \ref{fig:evaluation explain}. The criteria for computing the accuracy are summarized in the Methods section. We used accuracy as a metric to evaluate these two types of metrics. We also included the BLEU score as an additional metric. The methods we included in this section for benchmarking are LLaMA2-7B (LoRA), LLaMA2-7B (full), LLaMA2-7B (RAG), LLaMA2-13B (LoRA), LLaMA2-13B (full), Mistral-7B (LoRA), Mistral-7B (RAG), Gemma-7B (LoRA), Gemma-7B (RAG), LLaMAPro-8B (LoRA), LLaMAPro-8B (RAG), LLaMA3-8B (LoRA) and LLaMA3-8B (RAG). Here RAG represents the retrieval-augmented generation technology \citep{lewis2020retrieval}. We also considered including state-of-the-art models for general models and more biomedical-focused models, including GPT 3.5 \citep{brown2020language}, GPT 3.5 (RAG), GPT 4 \citep{achiam2023gpt}, BioMedLM \citep{BioMedLM2023}, bioGPT \citep{luo2022biogpt}, GeneGPT \citep{jin2023genegpt}, and BioMistral \citep{labrak2024biomistral}. Details of baselines are included in Appendix \ref{appendix:explain baseline}. Our evaluations are shown in Table \ref{tab:benchmark training}. 

\begin{table}[htbp]
  \centering
  \resizebox{\linewidth}{!}{
    \begin{tabular}{lrrrrrl}
    \toprule
    Model & \multicolumn{1}{l}{Factual Score} & \multicolumn{1}{l}{Structural Score} & \multicolumn{1}{l}{Average} & \multicolumn{1}{l}{BLEU} & \multicolumn{1}{l}{ROUGE1} & Finetuning? \\
    \midrule
    LLaMA2-7B (LoRA) & 0.850 & 1.000 & 0.925 & \textbf{0.395} & 0.580 & Y \\
    LLaMA2-7B (full) & 0.900 & 0.900 & 0.900 & 0.345 & 0.546 & Y \\
    LLaMA2-7B (RAG) & 0.650 & 0.800 & 0.725 & 0.155 & 0.324 & N \\
    LLaMA2-13B (LoRA) & 0.900 & 1.000 & 0.950 & 0.374 & \textbf{0.600} & Y \\
    LLaMA2-13B (RAG) & 0.600 & 1.000 & 0.800 & 0.196 & 0.358 & N \\
    \textbf{Mistral-7B (LoRA)} & 0.950 & 1.000 & \textbf{0.975} & 0.349 & 0.592 & Y \\
    Mistral-7B (RAG) & 0.650 & 0.050 & 0.350 & 0.127 & 0.269 & N \\
    \textbf{Gemma-7B (LoRA)} & 0.950 & 1.000 & \textbf{0.975} & \textbf{0.425} & \textbf{0.613} & Y \\
    Gemma-7B (RAG) & 0.350 & 1.000 & 0.675 & 0.073 & 0.183 & N \\
    \textbf{LLaMAPro-8B (LoRA) } & 0.950 & 1.000 & \textbf{0.975} & \textbf{0.387} & \textbf{0.627} & Y \\
    LLaMAPro-8B (RAG) & 0.000 & 0.000 & 0.000 & 0.000 & 0.000 & N \\
    LLaMA3-8B (LoRA) & 0.900 & 1.000 & 0.950 & 0.324 & 0.541 & Y \\
    LLaMA3-8B (RAG) & 0.250 & 0.400 & 0.325 & 0.011 & 0.196 & N \\
    GPT 3.5 & 0.900 & 1.000 & 0.950 & 0.123 & 0.306 & N \\
    GPT 3.5 (RAG) & 0.850 & 1.000 & 0.925 & 0.144 & 0.312 & N \\
    GPT 4 & 0.850 & 1.000 & 0.925 & 0.118 & 0.309 & N \\
    BioMedLM & 0.000 & 0.000 & 0.000 & 0.046 & 0.193 & N \\
    bioGPT & 0.000 & 0.000 & 0.000 & 0.041 & 0.186 & N \\
    GeneGPT & 0.750 & 1.000 & 0.875 & 0.141 & 0.304 & N \\
    BioMistral & 0.000 & 0.000 & 0.000 & 0.000 & 0.000 & N \\
    \bottomrule
    \bottomrule
    \end{tabular}%
    \vspace{-0.3cm}
    }
  \caption{Evaluations for the task: Gene Function Description. We boldfaced the top three methods ranked by Average scores, BLEU scores and ROUGE1 scores.}
  \label{tab:benchmark training}%
\end{table}%
Based on our evaluation results, we found that \method{} based on finetuning Mistral-7B, Gemma-7B, and LLaMAPro achieved the best results by averaging the scores from both content level and structure level. The latter two models also had better performances evaluated by BLEU and ROUGE1 scores. We computed the Pearson correlation between BLEU (ROUGE1) scores and average scores, which showed significant positive correlations (coefficient=0.76, p-value=9.5e-5 for BLEU, and coefficient=0.80, p-value=1.6e-5 for ROUGE1), which lends support of our proposed metric. We note that finetuned Mistral-7B outperformed finetuned models from the LLaMA family with different scales, but also outperformed large-scale general LLMs including GPT 3.5 and GPT 4. On the other hand, other methods including bioGPT and BioMistral did not generate meaningful descriptions for gene functions. Meanwhile, methods based on RAG also did not satisfy our requirements and the generated outputs sometimes have multiple paragraphs. Therefore, the structure score of Mistral-7B (RAG) is low. Our conclusions also agree with recent research \citep{wu2024faithful} challenging the capacity of RAG in addressing biomedical problems. Therefore, we still need to finetune models for domain-specific tasks.

Moreover, the model outputs under different cases all failed in summarizing the numerical information of genes including Ensemble id \citep{howe2021ensembl} and HUGO gene nomenclature committee (HGNC) id \citep{povey2001hugo}. To tackle this problem, we have developed a retrieval database to replace the incorrect information of model output as an important post-processing step. By adding this component, our generated summary is precise in describing both the functional and identifiable information of genes. The details of the post-processing method are included in the Methods section. The outputs of different models for evaluations are in Supplementary File 1.

\subsection{Leveraging the diverse of datasets to finetune a novel LLM}
As shown in Figure \ref{fig:model overview}, we also used the outputs generated by GPT-3.5 to enrich the data diversity for model training. To ensure the correctness of training datasets, we only focused on protein-encoding genes for the outputs of GPT 3.5. While for non-coding genes and pseudogenes, we only used the descriptions from NCBI. There are two advantages brought by such a data augmentation approach. The first advantage is that GPT 3.5 has specific knowledge of certain genes, thus helping our AI assistant answer questions related to genes that are not collected in the NCBI database. The other advantage is that GPT 3.5's output is more like human language and has more comprehensive functional information to enhance the quality of the model output. A comparison of outputs for models trained with different datasets is shown in Extended Data Figure \ref{fig:llm example}. From this figure, we can find that introducing more data can help LLMs generate more informative summaries for the gene GLI1 by reducing redundant information. Details of model outputs are summarized in Supplementary File 1.

\begin{figure*}[htbp]
    \centering
    \includegraphics[width=0.8\textwidth]{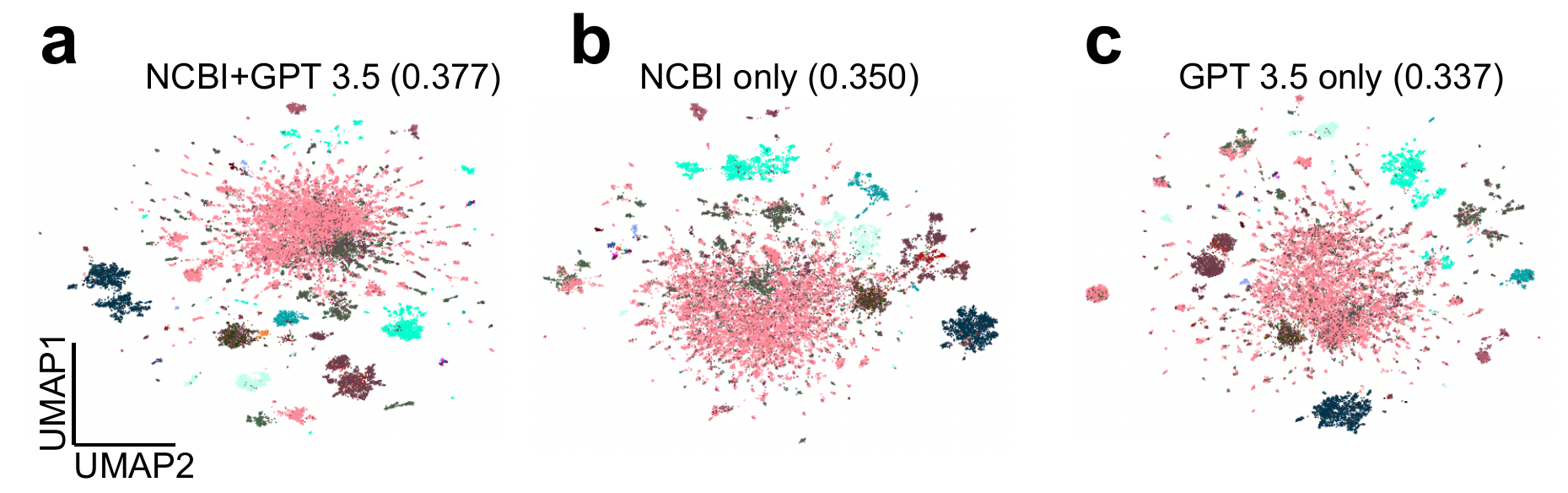}
    \caption{UMAPs for the gene embeddings colored by gene functional information. Panels (a)-(c) represent the outputs of LLMs trained based on datasets from NCBI+GPT 3.5, NCBI only and GPT 3.5. We report the NMI score of each embeddings followed by their sources.
    \label{fig:figure gene embeddings}}
    \vspace{-0.3cm}
\end{figure*}

To visualize the outputs of \method{} after finetuning, we utilize the embeddings layer from OpenAI to transfer the descriptions of genes into numerical vectors (as gene embeddings) and visualize the gene embeddings using Uniform Manifold Approximation and Projection (UMAP) \citep{mcinnes2018umap}. In Figure \ref{fig:figure gene embeddings} (a), we label the genes using gene functional information from \citep{theodoris2023transfer}. We can find that genes with similar functions display shared patterns. Details of labels are summarized in Extended Data Figure \ref{supfig: annotat gene}. Furthermore, the Gene Ontology Enrichment Analysis (GOEA) \citep{ashburner2000gene, gene2023gene, fang2023gseapy} results are summarized in Extended Data Figure \ref{fig:figure pathway data}, which include top 10 pathways ranked by $-log(\text{Adjusted P-value})$ (P-value corrected by Bonferroni method) for the top 10 clusters ranked by the number of genes. We find many pathways related to important biological activities such as RNA transcription and metabolism. Therefore, our sentences cover the similarities and differences of the genes' functions. 

\subsection{Finetuning a MLLM for genomic and proteomic application}
By incorporating the image information into the model training process, we can construct domain-specific MLLMs for several downstream tasks. We designed two tasks and demonstrated the potential of MLLMs for analyzing proteins and spatial transcriptomics by finetuning LLaVA on these two novel tasks. The first task is known as protein classification, which means we intend to identify the specific protein with its image. The images of protein structure for training and testing come from the databases of AlphaFold \citep{jumper2021highly}. The second task is known as spatial marker gene identification, which means that for a gene $g$, we intend to know whether it is a maker gene of cell type $c$ based on the spatial transcriptomic data. The spatial transcriptomic data for training and testing come from \citep{lin2020scclassify} collected from human breast tissue. Both tasks have not been discussed previously and are important for biomedical research. As for baselines, we included LLaVA-7B, MoE-LLaVA, GPT-4, and Gemini. Our finetuned version of LLaVA is known as LLaVA-7B (LoRA). Details of baselines are included in Appendix \ref{appendix:explain baseline}.

For the first task, we compared the three baseline models with our finetuned model, and the results are summarized in Table \ref{tab:protein data}. Based on this table, we found that the baselines without finetuning cannot generate correct labels for proteins from images. Moreover, the output of GPT-4 even does not follow the structure requirements from our prompts, which questioned the contributions of large-scale models for this task. However, if we finetuned LLaVA based on the training dataset, the factual score increases while the high structural score is also preserved. Therefore, leveraging domain-specific knowledge is an important step to handle this classification problem. However, only partial protein names were matched for every sample, thus the finetuning process still needs further investigation for performance improvement. 

\begin{table}[htbp]
  \centering
  \resizebox{\linewidth}{!}{
    \begin{tabular}{lrrrl}
    \toprule
    Model & \multicolumn{1}{l}{Factual score} & \multicolumn{1}{l}{Structural score} & \multicolumn{1}{l}{Average} & Finetuning? \\
    \midrule
    LLaVA-7B & 0     & 1     & 0.5   & N \\
    MoE-LLaVA & 0     & 1     & 0.5   & N \\
    GPT-4v & 0     & 0     & 0     & N \\
    GPT-4o & 0     & 0     & 0     & N \\
    Gemini & 0     & 0     & 0     & N \\
    \textbf{LLaVA-7B (LoRA)} & 0.29 & 1     & 0.645 & Y \\
    \bottomrule
    \bottomrule
    \end{tabular}%
    }
  \caption{Evaluations for the task: Protein Classification.}
  \label{tab:protein data}%
  \vspace{-0.3cm}
\end{table}%

For the second task, we compared the three baseline models with our finetuned model, and the results are summarized in Table \ref{tab:spatial marker gene ident}. According to this table, the results from GPT-4 are slightly better than random guesses, and all the baseline models did not show a strong ability to identify marker genes for different cell types given the spatial figures colored by gene expression levels. However, LLaVA-7B and GPT-4 did not perform well in following the structural requirements. Moreover, prior knowledge in the training process might help in this prediction process. For example, all the methods predicted gene CD4 as a marker gene of CD4 T cell, which matched the biological knowledge, and the reason is from the existing information of the pre-training text databases. Moreover, our finetuned LLaVA outperformed other models again in the factual scores, which had 90\% accuracy. In addition, the outputs of our finetuned LLaVA also followed the requirements of structure existing in the prompts. Therefore, finetuning an MLLM with image information could boost the ability of models to learn about specific biological problems, especially for tasks related to spatial transcriptomics. 

\begin{table}[htbp]
  \centering
  \resizebox{\linewidth}{!}{
    \begin{tabular}{lrrrl}
    \toprule
    Model & \multicolumn{1}{l}{Factual Score} & \multicolumn{1}{l}{Structural Score} & \multicolumn{1}{l}{Average} & Finetuning? \\
    \midrule
    LLaVA-7B & 0.5     & 0.425     & 0.4625   & N \\
    MoE-LLaVA & 0.5     & 1     & 0.75   & N \\
    GPT-4v & 0.575     & 0.125     & 0.35     & N \\
    GPT-4o & 0     & 0     & 0     & N \\
    Gemini & 0.6     & 0.025     & 0.3125 & N \\
    \textbf{LLaVA-7B (LoRA)} & 0.9 & 1     & 0.95 & Y \\
    \bottomrule
    \bottomrule
    \end{tabular}%
    }
  \caption{Evaluations for the task: Marker genes.}
  \label{tab:spatial marker gene ident}%
  \vspace{-0.3cm}
\end{table}%

\subsection{Sensitivity analysis}
\begin{figure*}[htbp]
    \centering
    \includegraphics[width=0.7\textwidth]{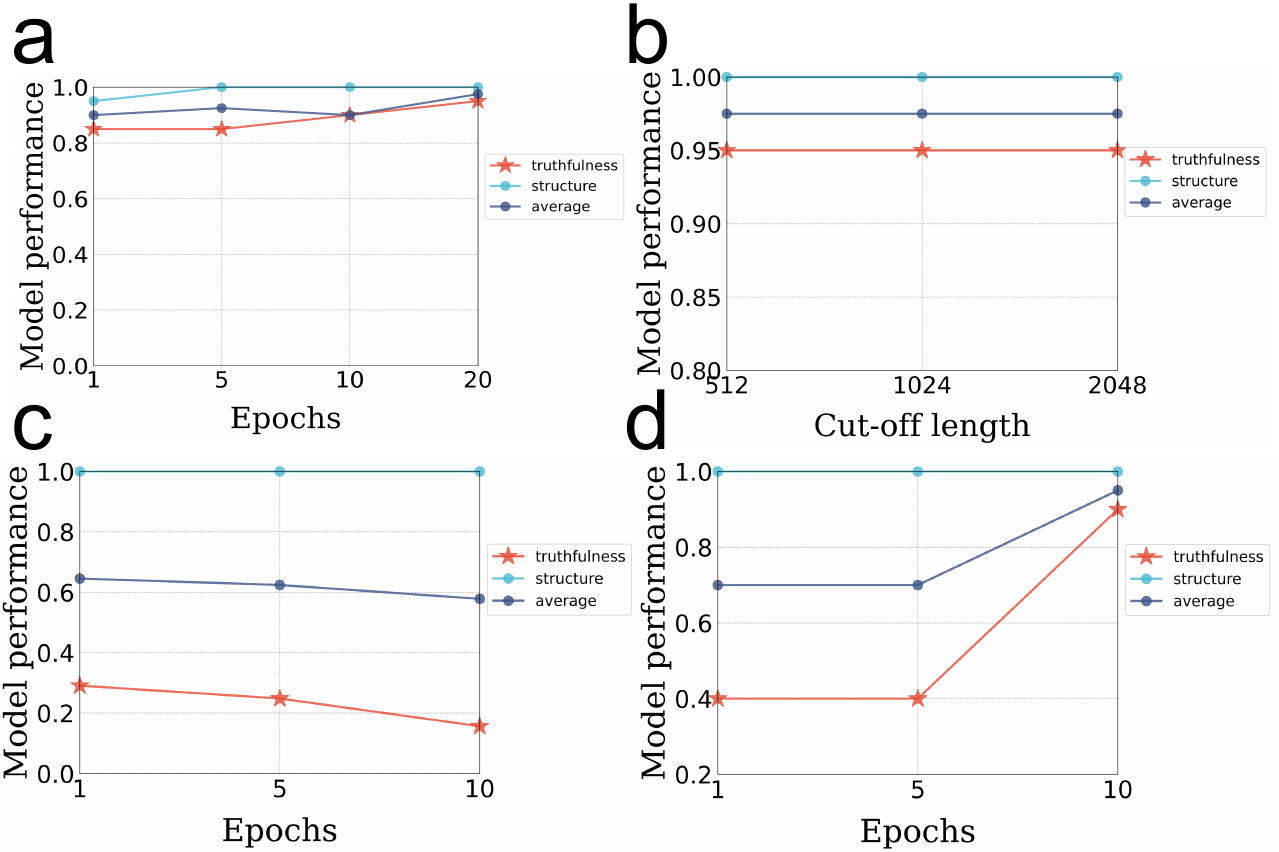}
    \caption{Results of sensitivity analysis for the training of different models. (a) The relation between the number of epochs and model performance of LLMs. (b) The relation between the number of cut-off length and model performance of LLMs. (c) The relation between the number of epochs and model performance of MLLMs for the protein task. (d) The relation between the number of epochs and model performance of MLLMs for the gene task.}
    \label{fig:llm sensi analysis}
    \vspace{-0.3cm}
\end{figure*}

In this section, we analyzed the factors that may affect the finetuning process, including hyper-parameters and data ablation. The results are summarized in Figure \ref{fig:llm sensi analysis}. We utilized default hyper parameters to finetune different LLMs and MLLMs, and selected the best base model used in this section for sensitivity analysis.

We first discuss the sensitivity of LLM training. For hyper-parameter tuning, we considered tuning the number of epochs and cut-off length (it means we use different lengths to truncate one sentence). For data ablation, we considered three conditions: 1. only NCBI data (NCBI), 2. only GPT 3.5 data (GPT 3.5), and 3. the combination of NCBI and GPT data (NCBI$+$GPT 3.5). Extended Data Figure \ref{supfig:ncbi gpt data abla} shows the results of data ablation, which suggests that integrating both NCBI and GPT 3.5 leads to the best performance of LLM in this task. The LLM based on GPT 3.5 to finetune does not accurately summarise the function of non-protein-encoding genes, as shown in Supplementary File 1. Moreover, LLMs trained based on the combination of datasets also generated a better representation. Comparing with outputs of LLMs trained based on the second condition (shown in Figure \ref{fig:figure gene embeddings} (b)) and the third condition (shown in Figure \ref{fig:figure gene embeddings} (c)), embeddings from our current design also have the highest NMI score. Figure \ref{fig:llm sensi analysis} (a) shows the relation between epochs and model performance, which suggests that more epochs lead to better model performance in the description generation task. Figure \ref{fig:llm sensi analysis} (b) shows the relation between cut-off length and model performance, which suggests that cut-off length does not affect the performance of LLM in this task. 

We then discuss the sensitivity of MLLM training. Based on our observations in the training process of LLMs, we only considered hyper-parameter tuning of the number of epochs. The relation between the number of epochs and model performance is illustrated in Figures \ref{fig:llm sensi analysis} (c) and (d) for the protein task and in Figure for the gene task. According to Figure \ref{fig:llm sensi analysis} (c), increasing the number of epochs leads to a model performance drop in the training process of MLLMs for the protein task, reflected in the average score and the truthful score. However, according to Figure \ref{fig:llm sensi analysis} (d), increasing the number of epochs leads to better model performance in the training process of MLLMs for the gene task, reflected in the average score and the score for truthfulness. Therefore, task variability affects the relationship between hyper-parameters and model performance, possibly due to the complexity of biological data and questions.

We also illustrate the relation between top-k candidates (It means the number of top samples retrieved from the reference data ranked by vector similarity) and model performance for RAG-based LLMs in Extended Data Figure \ref{supfig:rag sup result}. This figure shows that increasing top-k candidates leads to a model performance drop, especially for the score of structure. The RAG-based LLM using the largest top-k generated results free of structure, which was not suitable for the application of a useful AI assistant. Moreover, we considered using Lora$+$ \citep{hayou2024loraplus} to replace LoRA in the training process, but the average score did not change. The running time of the finetuning process is summarized in Supplementary File 2. 

\section{Discussion}
Foundation models, especially LLMs, are powerful tools for scientific research. In this manuscript, we introduced \method{}, a collection of LLMs and MLLMs tailored for genomic and proteomic tasks. We compared different strategies and base models for generating accurate function descriptions of genes based on the adapted LLMs. Moreover, we also explored the applications of MLLMs for a protein-level task and a gene-level task, as an example of leveraging the multimodal information. The models in \method{} demonstrated better performance in the selected tasks than closed-source models, thus advocating the contributions of open-source base models for scientific research. We also discussed the factors affecting model adaptation, which advanced this field by offering suggestions.

Our future work aims to increase the generation quality of outputs from \method{} with more advanced techniques. We will also investigate why RAG-based LLMs fail and improve them. Moreover, we will collaborate with other institutes to access more high-quality datasets for model adaptation. The next-generation multimodal AI assistant for genomics and proteomics will come soon.

\section{Limitations}

First, finetuning an FM or LLM requires a large amount of computing resources. Based on our testing results, we need at least one H100 or equivalent GPU cores to perform finetuning with LoRA. Secondly, to utilize methods based on closed-source large models, including GPT series and GeneGPT, we need to use the OpenAI API to call the method, thus we need additional resources for deploying such a method. Finally, a good domain-specific FM should align with the version of state-of-the-art base models. Therefore, we should keep our model updated by aligning the best version of \method{} with the best open-source base model. Moreover, due to the limitation of computing resources, we only considered 7B-level and 13B-level models in the current version. In the future, we need to explore LLMs or MLLMs with larger scales. Finally, our evaluations for the domain-specific FMs only focus on three tasks and more tasks should be included in the future.

\section{Ethics Statement}
The users are solely responsible for the content they generate with models in \method{}, and there are no mechanisms in place for addressing harmful, unfaithful, biased, and toxic content disclosure. Any modifications of the models should be released under different version numbers to keep track of the original models related to this manuscript. 

The target of current \method{} only serves for academic research. The users cannot use it for other purposes. Finally, we are not responsible for any effects of the use of the model.

\section{Codes and Reproductivities}
The codes we used in this manuscript can be found in \url{https://github.com/HelloWorldLTY/Geneverse}. We used the MIT licenses. All of the model weights can be found in the Huggingface Space. We will release our training datasets after peer review. We used two NVIDIA H100 GPU cores for model training, while the maximum system memory for training is 100 GB. 

\section{Contribution}
 T.L. proposed this study. T.L., Y.X. and X.L. designed the model. T.L. ran all the experiments. T.L., X.L. and H.Z. wrote the manuscript. H.X., W.Z., and H.Z. supervised this study.

\bibliography{main}

\counterwithin{figure}{section}
\renewcommand{\figurename}{Extended Data Fig.}
\renewcommand\thefigure{\arabic{figure}} 

\newpage
\appendix
\input{Appendix}

\end{document}

%% file: Appendix.tex
\section{Explanations of baseline models.}
\label{appendix:explain baseline}
Here we highlight the summary of different selected open-source LLMs and MLLMs used in this manuscript for either baselines or base models.

For LLMs, we have the following models:

\begin{itemize}
    \item LLaMA2: LLaMA2 is a collection of pre-trained and finetuned LLMs for chatting and other downstream tasks. The base model we choose for finetuning is from LLaMA2-Chat, which is optimized for dialogue use cases.

    \item LLaMA3: LLaMA3 is a collection of pre-trained and finetuned LLMs for chatting and other downstream tasks. The base model we choose for finetuning is from LLaMA3-Instruct, which is optimized for dialogue use cases.

    \item Mistral: Mistral is an open-source LLM that is based on a different framework compared with LLaMA2. It can also be finetuned into different cases and we also use its chat mode as the base model.

    \item Gemma: Gemma is also an open-source LLM which is the state-of-the-art 7B-base model. It can also be finetuned into different cases and we also use its chat mode as the base model.

    \item LLaMAPro: LLaMAPro is also an open-source LLM modified based on LLaMA2. It offers extra blocks for finetuning and thus this approach can reduce the catastrophic forgetting of a finetuned LLM.

    \item GPT 3.5: GPT 3.5 is a generative pretraining model finetuned for the chatting task. We used the OpenAI API to access this model.

    \item GPT 4: GPT 4 is a generative pretraining model finetuned for the chatting task. We used the OpenAI API to access this model. GPT 4 also has the ability to handle image data as input.

    \item bioGPT: bioGPT is a finetuned GPT-2 model based on medical data. It is designed for the applications of LLMs in biomedical cases. 

    \item BioMedLM: BioMedLM is a finetuned GPT-2 model based on PubMed data. It is designed for the applications of LLMs in biomedical cases, especially for question-answering (QA) problems.

    \item GeneGPT: GeneGPT is a tool-augmented method for tasks related to medical queries. GeneGPT is based on searching documents to enrich the information in the prompts. We use the OpenAI API to access the base model of this method.

    \item bioMistral: bioMistral is a finetuned Mistral model based on medical data. It is designed for the applications of LLMs in biomedical cases. 
\end{itemize}

For MLLMs, we have the following models:

\begin{itemize}
    \item LLaVA: LLaVA is a collection of pre-trained and finetuned MLLMs for chatting and other downstream tasks. The base model we choose for finetuning is from LLaVA-v1.5, which is recommended in the tutorials of LLaVA.

    \item MoE-LLaVA: MoE-LLaVA is the mixture-of-experts (MOE) version LLaVA. It utilizes different experts in the fine-tuning stages with the same inputs from embeddings of images and embeddings of words. We access this model based on the Huggingface Space resource.

    \item Gemini: Gemini is a closed-source MLLM for various tasks. We use an interactive platform from Google to access this model.

    \item GPT-4: Introduced before. Here the GPT-4 model includes GPT-4v and GPT-4o, which represent their abilities to handle image-based data as prompts. GPT-4o is the most advanced model from OpenAI.
\end{itemize}

\clearpage
\section*{Supplementary figures}
This section contains all the supplementary figures, starting from next page.

\begin{figure*}[htbp]
    \centering
    \includegraphics[width=1.0\textwidth]{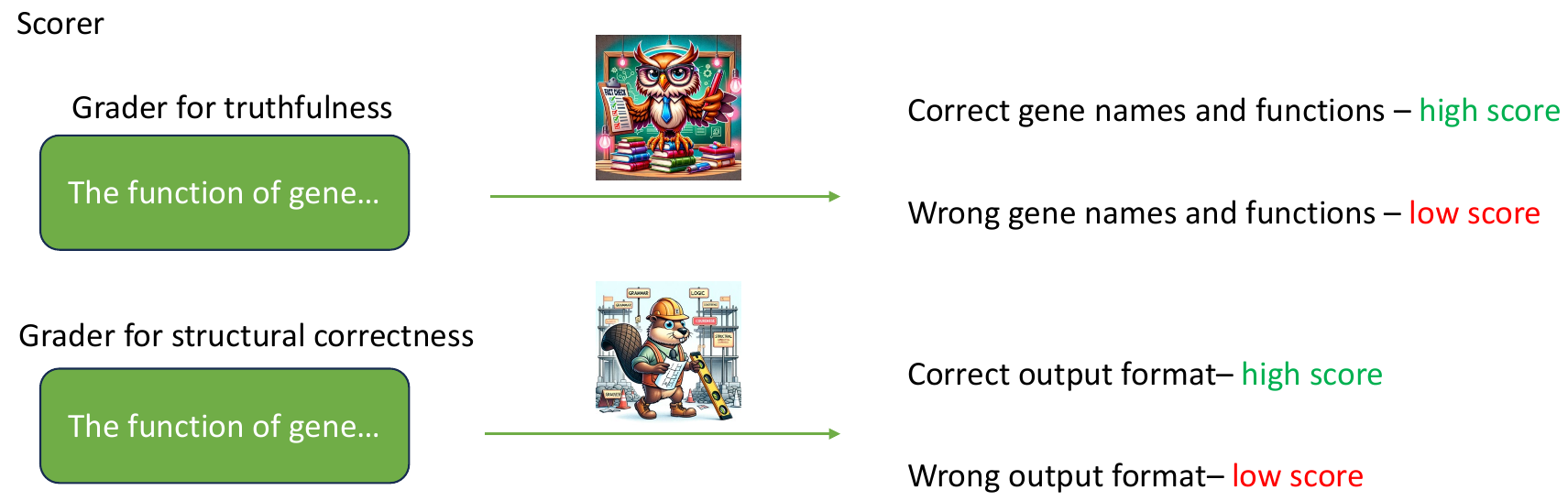}
    \caption{Definition of different evaluators or scorers. For the scorer focusing on truthfulness, we evaluate the matching level of model outputs for the description of gene properties and gene functions. For the scorer focusing on structural correctness, we evaluate the correctness of the structure of model outputs by comparing them to the limitations in the prompt. The logos of scorers are generated by DALLE \citep{dalle}.
    \label{fig:evaluation explain}}
\end{figure*}

\begin{figure*}[htbp]
    \centering
    \includegraphics[width=1.0\textwidth]{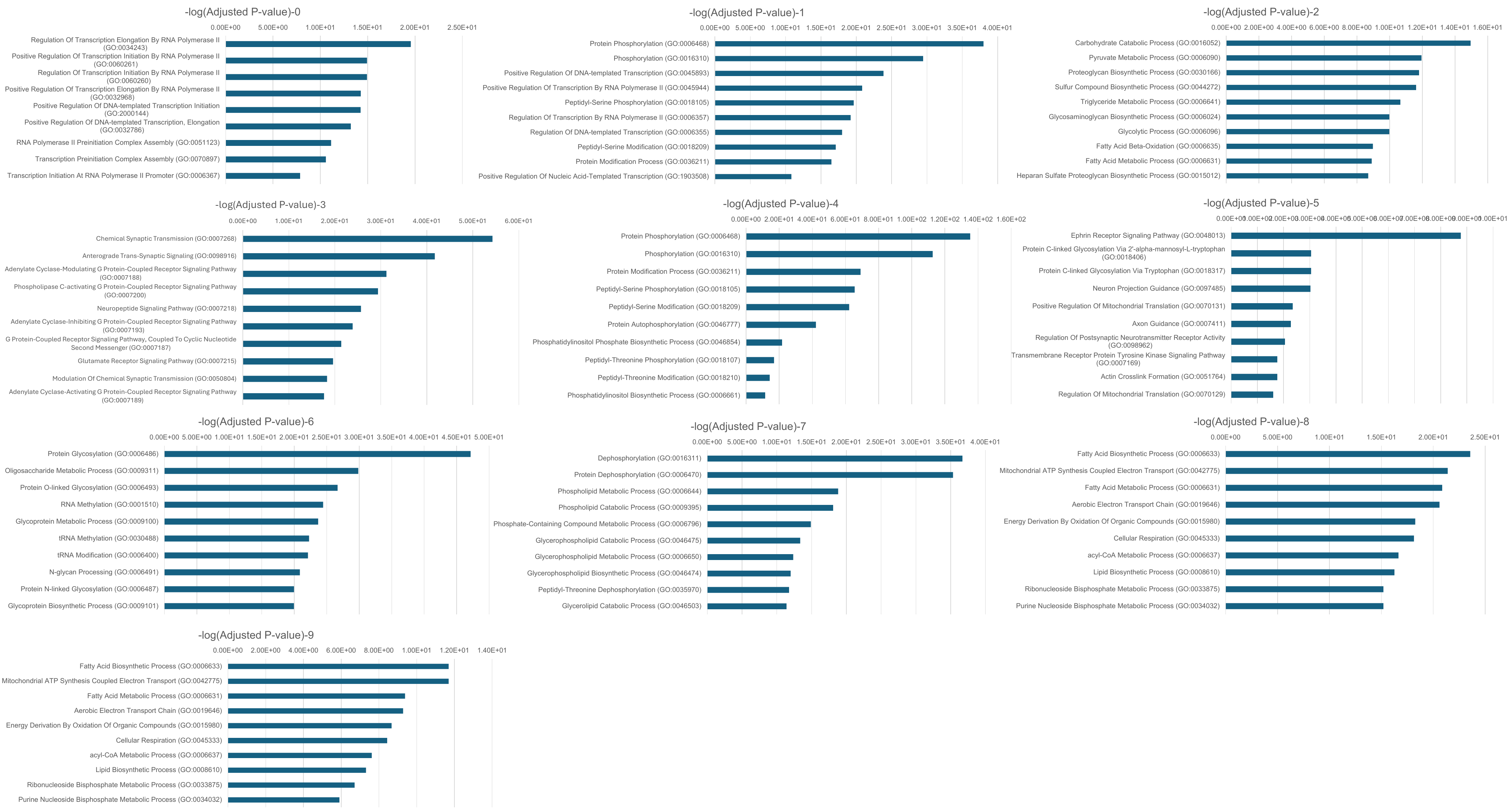}
    \caption{Figures of GOEA results. Each figure represents top 10 pathways in one cluster, and the pathways are ranked by $-\text{log(Adjusted P-value)}$.
    \label{fig:figure pathway data}}
\end{figure*}

\begin{figure*}[htbp]
    \centering
    \includegraphics[width=1.0\textwidth]{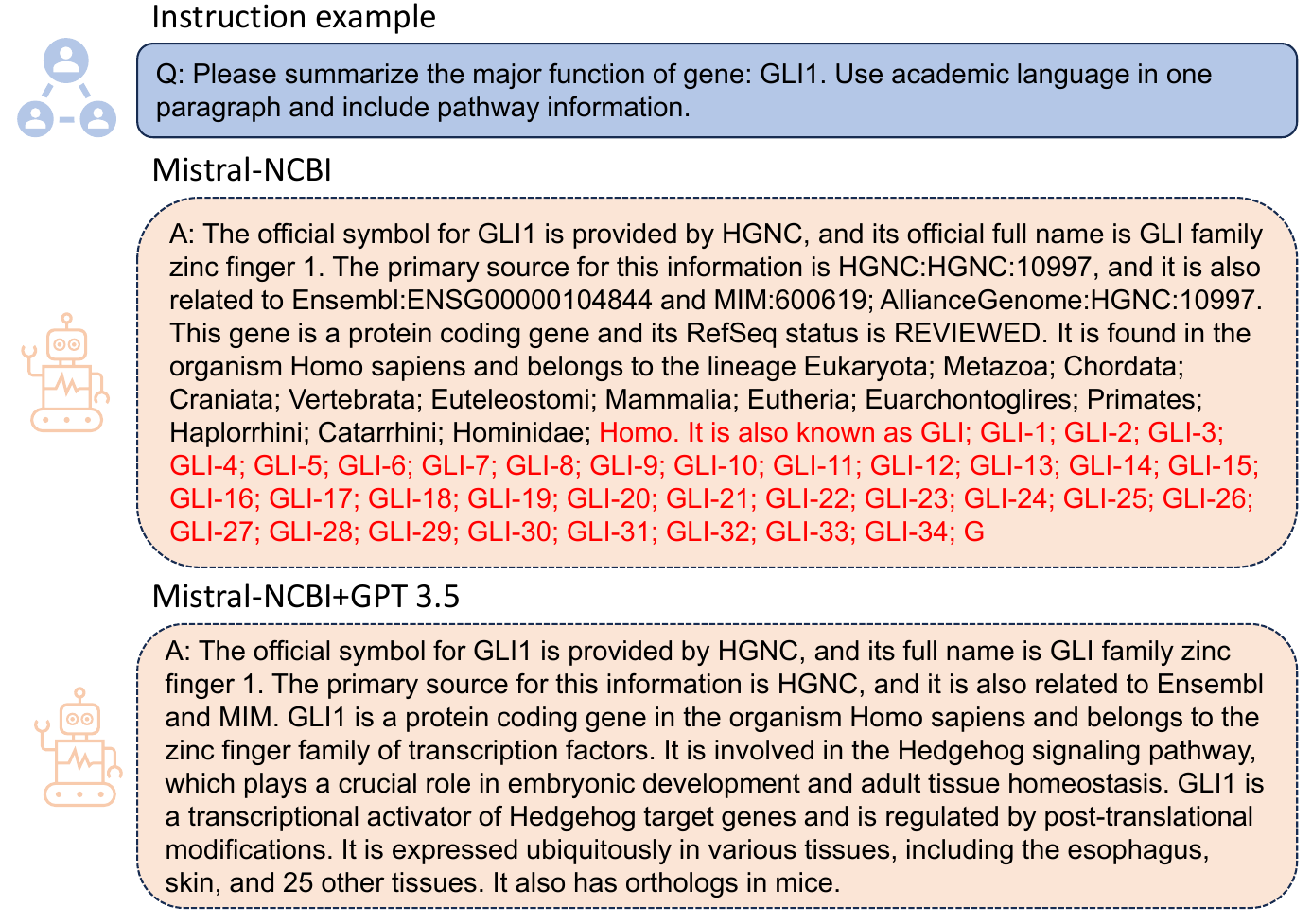}
    \caption{An example to illustrate the contribution of data augmentation. We compared the description of gene GLI1 from two models trained with different datasets but from the same instruction/prompt. We highlighted the redundant part in \textcolor{red}{red}.
    \label{fig:llm example}}
\end{figure*}

\begin{figure*}[htbp]
    \centering
    \includegraphics[width=1\linewidth]{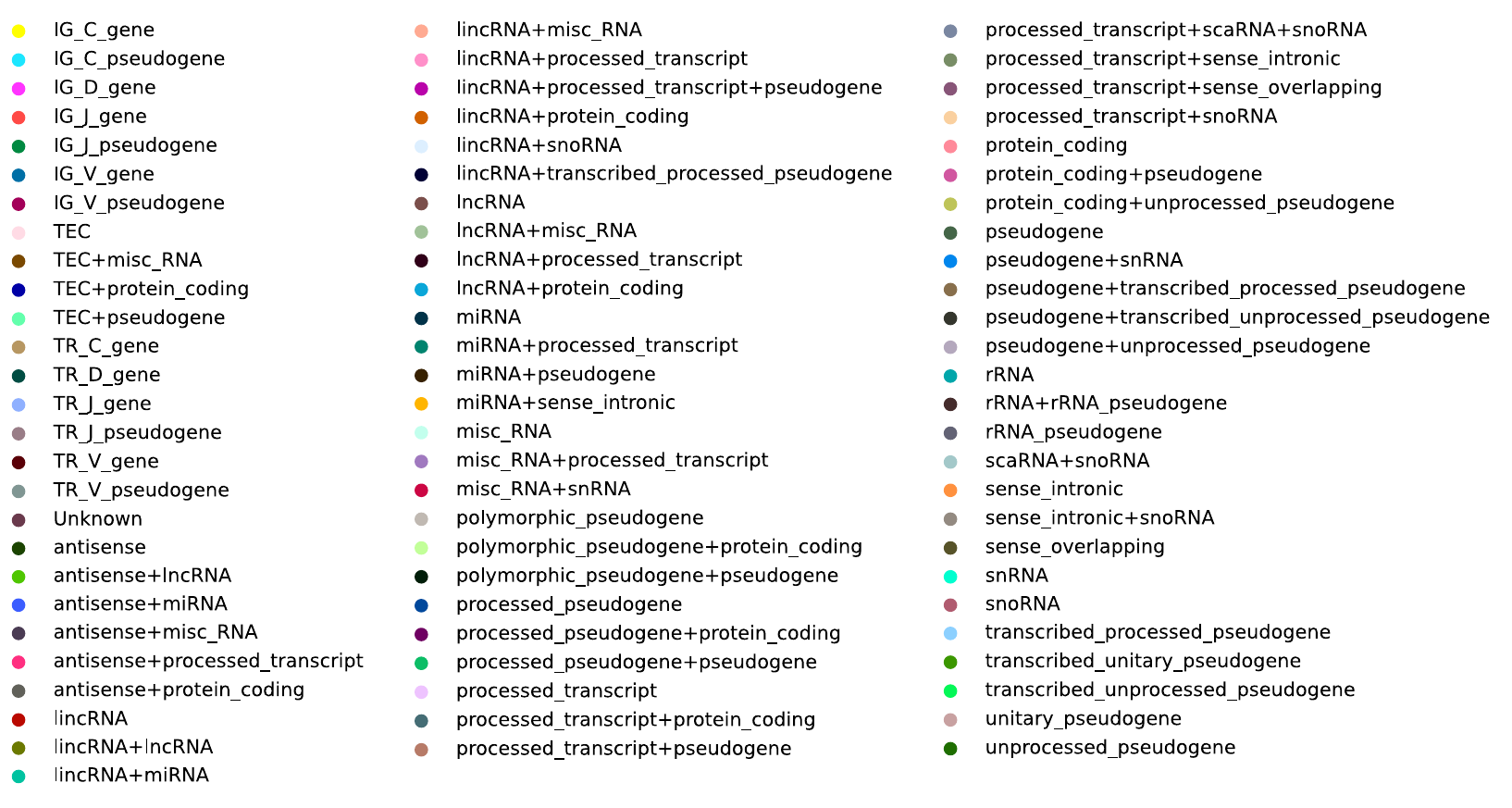}
    \caption{Annotation of gene functions by different colors.}
    \label{supfig: annotat gene}
\end{figure*}

\begin{figure*}[htbp]
    \centering
    \includegraphics[width=1.0\textwidth]{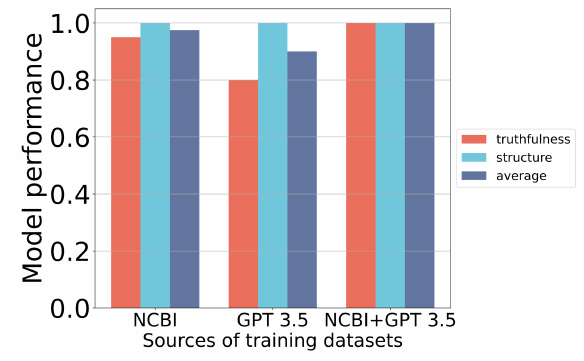}
    \caption{The results of data ablation for the training process.
    \label{supfig:ncbi gpt data abla}}
\end{figure*}

\begin{figure*}[htbp]
    \centering
    \includegraphics[width=1.0\textwidth]{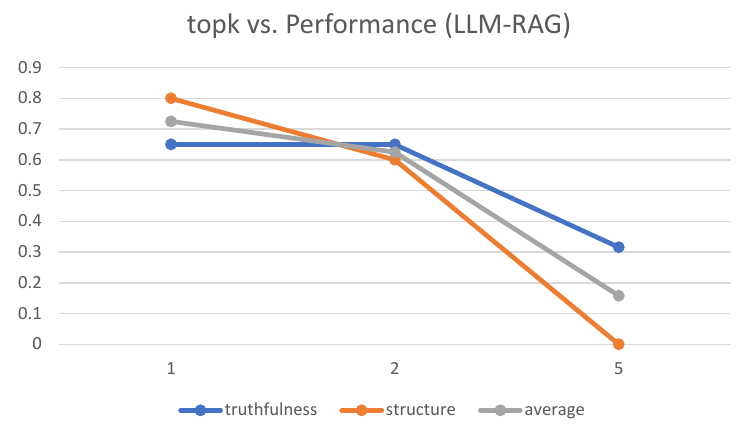}
    \caption{Relation between top-k candidates and model performance.
    \label{supfig:rag sup result}}
\end{figure*}